\documentclass{article}

\usepackage{arxiv}

\usepackage[utf8]{inputenc} 
\usepackage[T1]{fontenc}    
\usepackage{hyperref}       
\usepackage{url}            
\usepackage{booktabs}       
\usepackage{amsfonts}       
\usepackage{nicefrac}       
\usepackage{microtype}      
\usepackage{lipsum}
\usepackage{graphicx}
\usepackage{subfig}
\usepackage{amsmath}

\title{Designing morphologies of soft medical devices using cooperative neuro coevolution}

\author{
    Hugo Alcaraz-Herrera\\
    Unconventional Computing Laboratory, \\
    College of Arts, Technology and Environment, \\
    University of the West of England, \\
    Bristol, BS16 1QY, United Kingdom \\ \texttt{hugo.alcaraz@uwe.ac.uk}
\And
    Michail-Antisthenis Tsompanas  \\
    Unconventional Computing Laboratory \& \\
    School of Computing \& Creative Technologies,\\
    College of Arts, Technology and Environment, \\
    University of the West of England,\\ 
    Bristol, BS16 1QY, United Kingdom \\
    \texttt{antisthenis.tsompanas@uwe.ac.uk}
\And
       Andrew Adamatzky \\
       Unconventional Computing Laboratory,\\
    College of Arts, Technology and Environment, \\
    University of the West of England,\\ 
    Bristol, BS16 1QY, United Kingdom \\
\And
       Igor Balaz\\
       Laboratory for Meteorology, Physics and Biophysics,\\ Faculty of Agriculture, \\
       University of Novi Sad, \\ Trg Dositeja Obradovica 8, 21000, Novi Sad, Serbia
}

\begin{document}
\maketitle

\begin{abstract}
Soft robots have proven to outperform traditional robots in applications related to propagation in geometrically constrained environments.  Designing these robots and their controllers is an intricate task, since their building materials exhibit non-linear properties. Human designs may be biased; hence, alternative designing processes should be considered. We present a cooperative neuro coevolution approach to designing the morphologies of soft actuators and their controllers for applications in drug delivery apparatus. Morphologies and controllers are encoded as compositional pattern-producing networks evolved by Neuroevolution of Augmented Topologies (NEAT) and in cooperative coevolution methodology, taking into account different collaboration methods. Four collaboration methods are studied: $n$ best individuals, $n$ worst individuals, $n$ best and worst individuals, and $n$ random individuals. As a performance baseline, the results from the implementation of Age-Fitness Pareto Optimisation (AFPO) are considered. The metrics used are the maximum displacement in upward bending and the robustness of the devices in terms of applying to the same evolved morphology a diverse set of controllers. Results suggest that the cooperative neuro coevolution approach can produce more suitable morphologies for the intended devices than AFPO.
\end{abstract}


\section{Introduction}


Building soft robots is an intricate task because their materials present non-linear mechanical properties that are complex to characterise \cite{Hiller2014}. Consequently, finding an adequate design (i.e., morphology) implies considerable time and material resources since numerous prototype models must be tested in ``the real world'' \cite{Schulz2016}. Once an adequate soft robot model is found, the next step is designing a suitable controller. This process is typically as complex as finding a suitable morphology design due to the flexibility and ductility of materials. These features complicate the process of modelling the behaviour and movement of soft robots \cite{Wang2022}.


A promising strategy to address the inherent design challenges of soft robots is neuroevolution (NE). NE evolves the topology of artificial neural networks (ANNs) through a genetic algorithm (GA). One of the most popular NE-based approaches is Neuroevolution of Augmenting Topologies (NEAT) \cite{Stanley2002}, which has been previously implemented to generate robot morphologies \cite{Auerbach2011}, and control design \cite{Wen2017,Alcaraz2024controllers} in separate evolution runs. However, there is evidence that coevolution can perform better than evolution in diverse tasks such as feature weighting \cite{Blansch2005}. 


Thus, the primary objective of this research is to assess the suitability of NEAT under a cooperative coevolution \cite{Ma2019} scheme for designing soft actuator morphologies (SAMs) and their controllers. SAMs can be built using biological material (e.g., cells or tissue) with the purpose of enabling novel biohybrid machines to deliver drugs to areas of the human body that are difficult to reach. Thus the analysis here focuses on the capability of SAMs to make upward bending movements induced by the controllers evolving within the same methodology.

\section{Experimental setup}\label{sec:setup}

Under the scope of this research, three aspects need to be considered: the physics engine where SAMs are simulated, the specific configuration employed for NEAT during experimentation, and the general coevolutionary setup for experimentation. The following sections describe these aspects.

\subsection{Voxelyze}\label{sec:setup_voxelyze}


A physics engine called {\em Voxelyze} \cite{KriegmanGitHub} is used to simulate the mechanical response of SAMs. The minimum building block of this physics engine is a {\em voxel} and can represent different materials. 
In this research, two types of voxels are considered: active and passive.


Thus, Voxelyze acts as the fitness function of SAMs and their controllers. 
Furthermore, SAMs have a passive enclosure representing a bioreactor required for a muscle actuator with a nutrient supply (as conceptualized in \cite{Tsompanas2024}). The output of Voxelyze contains the trace of the free end (in the $x, y, z$ axes) of the simulated SAM. 
Finally, in terms of voxels, the dimensions of SAMs are as follows: 20 units on the $x$ axis and 8 units on the $y$ and $z$ axes. 

\subsection{NEAT configuration}\label{sec:setup_neat}

In this research {\em CPPN-NEAT} is implemented. This algorithm evolves a special type of ANNs called Compositional-Pattern Producing Networks (CPPNs) \cite{Stanley2007cppn}. The main feature of CPPNs is the capability of having, among others, periodic functions (e.g., sine, cosine) in hidden neurons. This allows the generation of symmetry and patterns in their topologies. Furthermore, a two-population cooperative coevolutionary approach is used. One population contains CPPNs encoding SAMs, and the other CPPNs encoding controllers. Although NEAT evolves both populations, each has a specific configuration for SAMs and controllers. 


Since SAMs are designed and simulated on a discrete three-dimensional canvas, it is compulsory to provide for each point $i$ across the canvas the presence (or not) of the voxel ($\nu_i$) and the type of material of the voxel ($m_i$). Thus, CPPNs encoding SAMs are queried as follows:

\begin{equation}\label{eq:setup_neat_catheter} 
    CPPN_{sam}(x_i,y_i,z_i) = \nu_i,m_i
\end{equation}

\noindent
where the tuple $(x_i,y_i,z_i)$ represents the coordinates of the $i$-$th$ point in the three-dimensional canvas. 
The two types of materials utilised here (as mentioned previously) are active and passive.






Due to SAMs being designed on a discrete three-dimensional canvas with two types of materials, the input of CPPNs encoding controllers contains the coordinates of the $i$ point across the canvas and the type of material of the voxel allocated at the $i$ point. Moreover, the output of CPPNs is the phase offset of the contraction of the active voxel allocated at the $i$ point across the canvas ($pho_i$). The phase offset represents the delay in the expansion behaviour of active voxels. Therefore, CPPNs are queried as follows:

\begin{equation}\label{eq:setup_neat_controller} 
    CPPN_{con}(x_i,y_i,z_i,m_i) = pho_i
\end{equation}

\noindent
where the tuple $(x_i,y_i,z_i)$ is the coordinates of the $i$-$th$ point in the three-dimensional canvas. 
Following insights from previous research \cite{Alcaraz2024controllers}, the output of CPPNs is clamped in the $[-2\pi,2\pi]$ range to provide complete sinusoidal-based contraction movements.

\subsection{Coevolutionary setup}\label{sec:setup_coevolutionary_setup}

The SAMs and controllers populations are composed of 25 individuals each. Under AFPO, the number of individuals is 50. Furthermore, for all approaches, each evolutionary run lasted 200 generations. Individuals (i.e., CPPNs) are initialised without hidden neurons, and the input neurons are fully connected to the output neurons with the minimal topology possible. 



Regarding the cooperative coevolutionary scheme between populations, they are evolved in a round-robin fashion (i.e., the same computational resources are assigned to each population) \cite{Ma2019}. Furthermore, four collaboration strategies are implemented: (a)~{\em n fittest individuals vs all} (NF) - The $n$ best individuals of one population are used to evaluate all the individuals of the other population; (b)~{\em n worst individuals vs all} (NW) - The $n$ worst individuals of one population are used to evaluate all the individuals of the other population; (c)~{\em n fittest and worst individuals vs all} (NFW) - The $n$ best and worst individuals of one population are used to evaluate all the individuals of the other population; and (d)~{\em n random individuals vs all} (NR) - Randomly, $n$ individuals are taken from one population to evaluate all individuals in the other population. Note that the range of values used for $n$ is $[1,2,3,5,10]$.



Finally, the experimental infrastructure detailed in \cite{Alcaraz2024locomotion,Alcaraz2024actuator} was replicated due to the significant computational time simulations required. Namely, the hardware used is as follows: {\em Processor}: ARM (virtualised), nine cores (18 threads), 3.20 GHz. {\em RAM Memory}: 16 GB, LPDDR5. Furthermore, NEAT was implemented under a client-server architecture to take advantage of distributed computation features.

\begin{figure*}[tb!]
  \centering
     \includegraphics[width=0.75\linewidth]{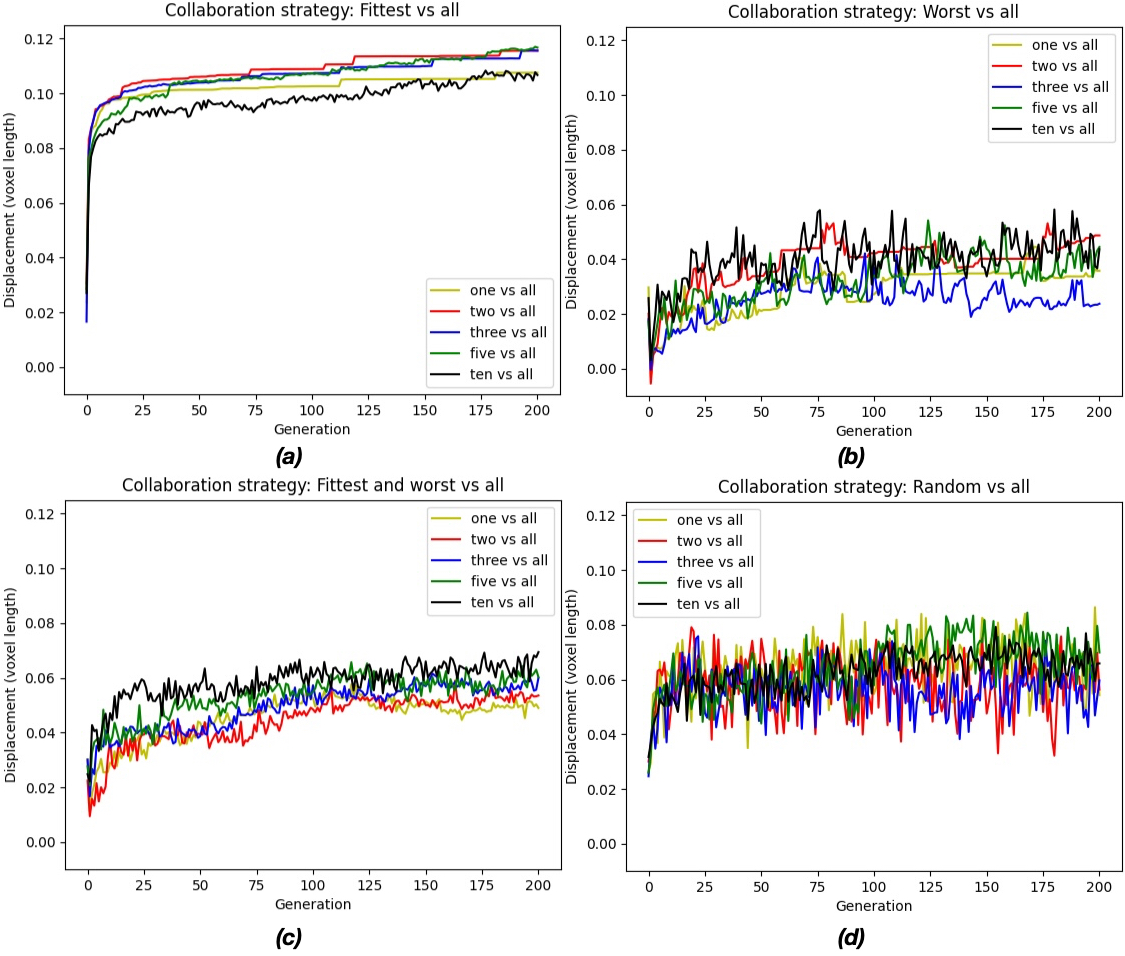}
  \caption{Mean performance observed of the fittest SAM using arithmetic mean under: (a) {\em n} fittest vs all; (b) {\em n} worst vs all; (c) {\em n} fittest and worst vs all; and (d) {\em n} random vs all.}
  \label{fig:experimental_arithmetic}
\end{figure*}


\section{Experimental results}\label{sec:experimental}

Two experiments are conducted to evaluate the performance of the system: (i) identifying the optimal cooperative configuration under four collaboration methods in terms of finding the SAM with the maximum displacement, 
and (ii) testing the robustness of the fittest SAMs discovered. 

\subsection{Optimal cooperative configuration}\label{sec:experimental_arithmetic}




The four collaboration strategies described in Section~\ref{sec:setup_coevolutionary_setup} are implemented in this experiment. For each collaboration strategy, individuals are evaluated using the arithmetic mean of the fitness of the corresponding collaborating individuals. Figure~\ref{fig:experimental_arithmetic} presents the mean performance of the fittest individual (i.e., the fittest SAM) across 10 evolutionary runs under four collaboration strategies.

When NF is utilised (Fig.~\ref{fig:experimental_arithmetic}-a), all approaches exhibit a clear tendency to evolve with scarce fluctuations. Whereas, when NW is used (Fig.~\ref{fig:experimental_arithmetic}-b), it is possible to distinguish traces of evolutionary behaviour among all approaches at the beginning. However, the evolutionary process stags from generation 60, exhibiting several fluctuations during evolution. Moreover, when NFW is implemented (Fig.~\ref{fig:experimental_arithmetic}-c), again, some traces of evolution are visible across all approaches until generation 90. Afterwards, evolution stags and numerous fluctuations can be observed. Finally, no tendency to evolve is observed under NR (Fig.~\ref{fig:experimental_arithmetic}-d), and significant fluctuations are exhibited.

Since variations NW, NFW, and NR did not reach the displacement threshold of 0.1, they are not considered for further analysis. Thus, the analysis is focused on the NF approach. First, all the data collected were tested and are not normally distributed (Shapiro-Wilk test; $p<0.05$). Then, employing Dunn's test, it is possible to confirm no significant differences between approaches $n=3$ and $n=5$ ($p>0.05$). However, significant differences exist among the rest of the approaches. Therefore, it is possible to rank the performance of the approaches: $n=2$ $>$ $n=3$ and $n=5$ $>$ $n=1$ $>$ $n=10$ ($p<0.05$).

These results indicate that NF generally performs suitably in finding adequate SAMs. Furthermore, employing two, three, or five fittest individuals in one population to evaluate all individuals in the other population obtains adequate results. However, using the two fittest individuals helps to enhance the evolutionary process. In contrast, the lack of the fittest individuals ($n=1$) or the excess of the fittest individuals ($n=10$) negatively impacts the evolutionary process.

\subsection{Multi-objective vs neuro coevolution}\label{sec:experimental_robustness}

SAMs are intended to be build in a real environment, where the controlling signals towards each part of the morphology may get distorted (i.e. cross-talk). Therefore, the operational environment from the controller point of view may be constantly changing. Thus, this experiment focuses on testing the robustness of the fittest morphology found within this study and the fittest morphology found by AFPO \cite{Alcaraz2024actuator}, in terms of controllers being applied.

The two SAMs are tested utilising 1000 different controller phase offsets, which, {\em a priori}, were randomly generated. Therefore, the robustness (i.e., aptitude) of SAMs is the mean displacement observed in 1000 simulations. 



\begin{figure}[tb!]
  \centering
     \includegraphics[width=0.75\linewidth]{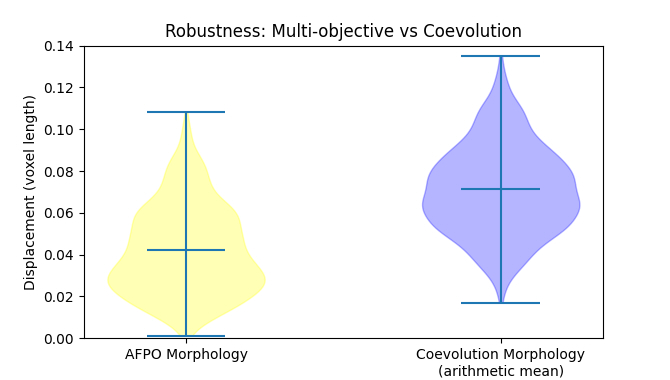}
  \caption{Displacement observed in the $yz$ plane under 1000 phase offset scenarios of the fittest SAM found under AFPO (left); coevolution (right).}
  \label{fig:experimental_robustness_II}
\end{figure}

Figure~\ref{fig:experimental_robustness_II} presents violin plots comparing the displacement observed in the $yz$ plane of the two SAMs found by AFPO and coevolution. Each violin plot shows the minimum, maximum, median and kernel density estimation of the frequency of values under 1000 different offset scenarios. 

There are significant differences in the performance of the SAMs. All the data collected were tested and are not normally distributed (Shapiro–Wilk test; $p<0.05$). Then, using Wilcoxon test and paired t-test, it is possible to confirm that significant differences exist between coevolution and AFPO-based morphology ($p<0.05$). 

Results suggest that coevolution can discover more suitable and robust SAMs than a multi-objective optimisation (i.e., AFPO). Arguably, the fact that controllers influence the evolutionary process of morphologies and vice versa allows coevolution to produce more bendable structures (and more accurate controllers) than the ones produced by the one-population-based mechanism of AFPO.

\begin{figure}[tb!]
  \centering
     \includegraphics[width=1.0\linewidth]{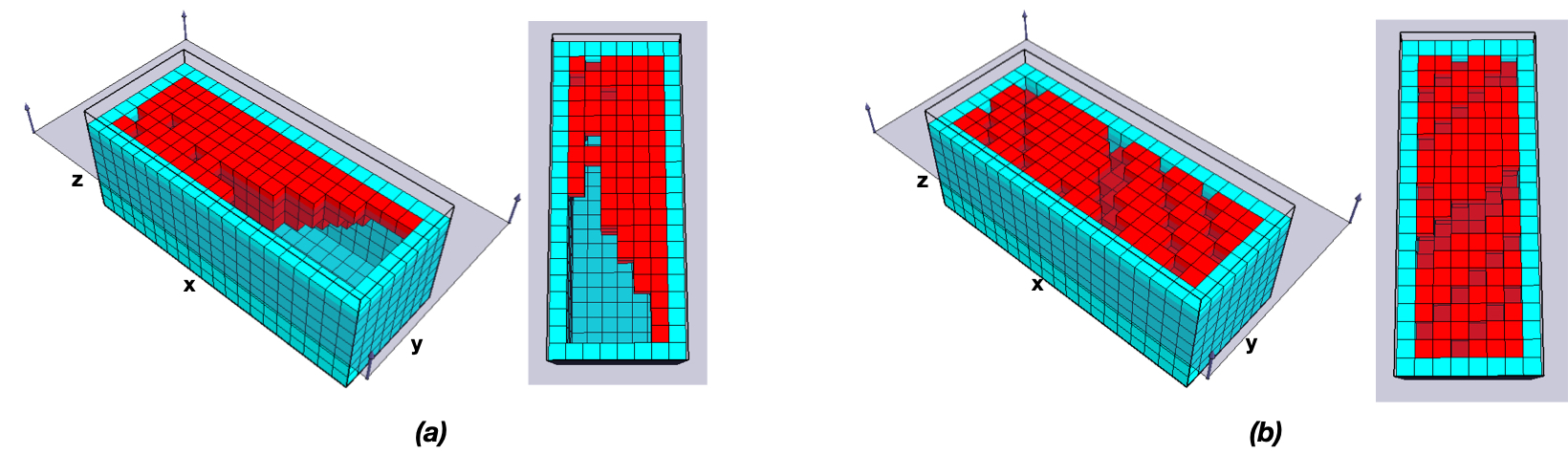}
  \caption{Fittest SAM found by: (a) AFPO and (b) coevolution. Red voxels represent active tissue, whereas blue voxels represent passive tissue.}
  \label{fig:experimental_robustness_morphologies_II}
\end{figure}

The fittest SAM discovered by AFPO (Fig.~\ref{fig:experimental_robustness_morphologies_II}-a) has a pyramidal-like pattern. At the top of the morphology, several active voxels are gradually decreasing towards the bottom of the morphology. The lack of active support at the bottom of the morphology possibly reduces its bending capability. Finally, the fittest SAM found by coevolution (Fig.~\ref{fig:experimental_robustness_morphologies_II}-b) exhibits a solid diagonal pattern of active voxels. This voxel allocation arguably allows the SAM to bend suitably.


\section{Conclusions}\label{sec:conclusions}

We studied the capacity of NEAT to generate adequate SAMs and controllers under a cooperative coevolutionary approach. The performance analysis focused on two metrics: (i)~finding the optimal configuration to discover suitable SAMs (and controllers) with the maximum displacement, and (ii)~testing the robustness of the fittest devices found. 

Regarding the first metric, the fitness value of individuals is based on the arithmetic mean. Four collaboration strategies were implemented: (a)~{\em n fittest individuals vs all}; (b)~{\em n worst individuals vs all}; (c)~{\em n fittest and worst individuals vs all}, and (d)~{\em n fittest and worst individuals vs all}. For each collaboration strategy, $n$ is assigned to each value of the $[1, 2, 3, 5, 10]$ range. 

Under the second metric, the fittest SAM found in the previous metric was compared against the fittest SAM found by AFPO, a popular multi-objective optimisation algorithm \cite{Schmidt2010}. The comparison consisted of testing the robustness of the two SAMs under 1000 different phase offset scenarios.

Results indicate that NEAT under a cooperative coevolutionary scheme, using {\em n fittest individuals vs all} as collaboration strategy, is more suitable to design SAMs than AFPO due to the following points: (i)~while evolving, controllers are interacting and hence, generating an ``evolutionary pressure'' over SAMs, and (ii)~the core mechanism is based on CPPNs, which helps NEAT to generate morphological patterns that contribute to the bending movement efficiency of SAMs and allows it to consider the morphological aspects and materials to design more accurate controllers.

Based on the results obtained from this research, future work directions are numerous. For instance, implementing under the same cooperative coevolutionary scheme Hypercube-based Neuroevolution of Augmenting Topologies (HyperNEAT) \cite{Stanley2009}. 
Another avenue of future work involves testing these NE-based approaches in more realistic scenarios and simulators, where significant environmental elements such as friction and viscosity can be included during simulations.





\section*{Acknowledgement}
This project has received funding from the European Union’s Horizon Europe research and innovation programme under grant agreement No. 101070328. UWE researchers were funded by the UK Research and Innovation grant No. 10044516.

\end{document}